%% file: main.tex
\documentclass[10pt,twocolumn,letterpaper]{article}

\usepackage{cvpr}

\usepackage{graphicx}
\usepackage{amsmath}
\usepackage{amssymb}
\usepackage{booktabs}

\input{preamble}

\definecolor{cvprblue}{rgb}{0.21,0.49,0.74}
\usepackage[pagebackref,breaklinks,colorlinks,citecolor=cvprblue]{hyperref}

\title{GaussianDiffusion: 3D Gaussian Splatting for Denoising Diffusion Probabilistic Models with Structured Noise}

\author{
Xinhai Li\textsuperscript{\rm 1}\\
Harbin Institute of Technology (Shenzhen)\\
Shenzhen, China\\
{\tt\small lixinhaiyk@gmail.com}
\and
Huaibin Wang\textsuperscript{\rm 2}\\
Harbin Institute of Technology (Shenzhen)\\
Shenzhen, China\\
{\tt\small 22S051022@stu.hit.edu.cn}
\and
Kuo-Kun Tseng\textsuperscript{\rm 3}\\
Harbin Institute of Technology (Shenzhen)\\
Shenzhen, China\\
{\tt\small kktseng@hit.edu.cn}
}

\begin{document}
\twocolumn[{%
\renewcommand\twocolumn[1][]{#1}%
\maketitle
\begin{center}
\centering
\end{center}%
\vspace{-2em}
\includegraphics[width=1\linewidth]{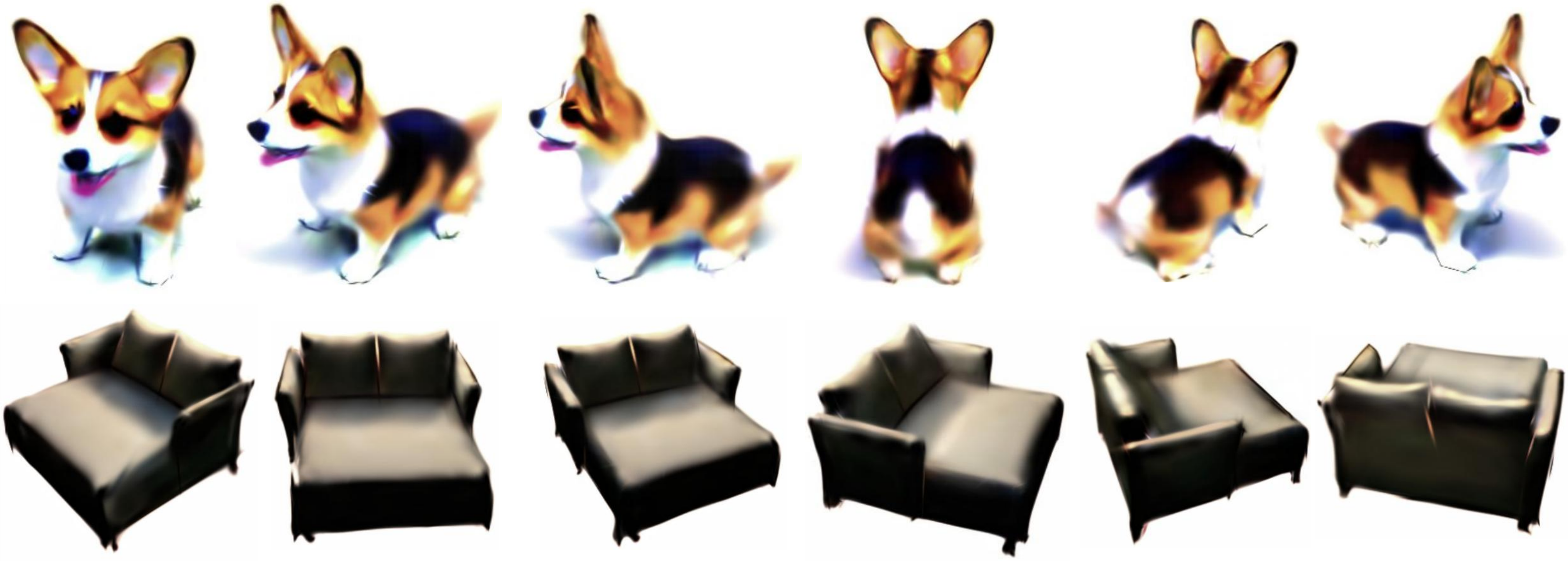}
    \captionof{figure}{GaussianDiffusion text-to-3D generation, given the text prompts 'a corgi' and 'a hamburger,'.}
    \vspace{1em}
}]

\input{sec/0_abstract}    
\input{sec/1_intro}
\input{sec/2_related}

\input{sec/3_method}

\input{sec/4_experiment}
\input{sec/5_conclusion}

{
    \small
    \bibliographystyle{ieeenat_fullname}
    \bibliography{main}
}


\end{document}

%% file: preamble.tex
\usepackage[dvipsnames]{xcolor}


%% file: sec/0_abstract.tex
\begin{abstract}
Text-to-3D, known for its efficient generation methods and expansive creative potential, has garnered significant attention in the AIGC domain. However, the pixel-wise rendering of NeRF and its ray marching light sampling constrain the rendering speed, impacting its utility in downstream industrial applications. Gaussian Splatting has recently shown a trend of replacing the traditional pointwise sampling technique commonly used in NeRF-based methodologies, and it is changing various aspects of 3D reconstruction. This paper introduces a novel text to 3D content generation framework, Gaussian Diffusion, based on Gaussian Splatting and produces more realistic renderings. The challenge of achieving multi-view consistency in 3D generation significantly impedes modeling complexity and accuracy. Taking inspiration from SJC, we explore employing multi-view noise distributions to perturb images generated by 3D Gaussian Splatting, aiming to rectify inconsistencies in multi-view geometry. We ingeniously devise an efficient method to generate noise that produces Gaussian noise from diverse viewpoints, all originating from a shared noise source. Furthermore, vanilla 3D Gaussian-based generation tends to trap models in local minima, causing artifacts like floaters, burrs, or proliferative elements. To mitigate these issues, we propose the variational Gaussian Splatting technique to enhance the quality and stability of 3D appearance. To our knowledge, our approach represents the first comprehensive utilization of Gaussian Diffusion across the entire spectrum of 3D content generation processes.
\end{abstract}

%% file: sec/1_intro.tex
\section{Introduction}

3D content generation~\cite{lin2023magic3d,poole2022dreamfusion,chen2023fantasia3d,tang2023make,qian2023magic123,shi2023mvdream,liu2023syncdreamer,tang2023dreamgaussian,seo2023let,liu2023one,shi2023zero123pp} based on large language models~\cite{radford2021learning,dosovitskiy2020image} has garnered widespread attention, marking a significant advancement in modern industries such as gaming, media, interactive virtual reality, and robotics applications. Traditional 3D asset creation is a labor-intensive process that relies on well-trained designers to produce a single 3D asset, involving tasks like geometric modeling, shape baking, UV mapping, material design, and texturing. Hence, there is a pressing need for an automated approach to achieve high-quality 3D content generation with consistent geometry across multi-views and rich materials and textures.

Gaussian Splatting~\cite{kerbl20233d} is currently showing a trend of replacing the traditional pointwise sampling approach used in NeRF-based methodologies, leading to a paradigm shift in multiple domains of 3D reconstruction. The unique representation offered by 3D Gaussian splatting not only enables a continuous portrayal of 3D scenes but also seamlessly integrates with traditional rendering pipelines in a discreet form. Therefore, it greatly accelerates the rendering speed of 3D models in downstream applications.

DreamFusion~\cite{poole2022dreamfusion} pioneered the learning of 3D contents from 2D diffusion models through score distillation sampling (SDS), and then followed these excellent text-to-3D solutions ~\cite{wang2023score,seo2023let,tang2023dreamgaussian,tsalicoglou2023textmesh,yu2023points,zhu2023hifa,li2023focaldreamer,wang2023prolificdreamer,huang2023dreamtime,chen2023fantasia3d}. SJC solves the out-of-distribution (OOD) problem between the standard normal of the 2D diffusion model input and the 3D rendering, through secondary noise addition to 3D rendering. 3DFuse~\cite{seo2023let} uses a multi-stage coarse-to-fine method to guide the direction of 3D model generation, uses view-specific depth as a condition, and applies controlNet~\cite{zhang2023adding} to control the direction of Diffusion model generation. 

However, there are currently several challenges associated with learning 3D content from a 2D pretrained diffusion model. When using a given text prompt, the generation of multi-view 2D diffusion tends to result in multi-view geometric consistency issue. Simultaneously, the rendering speed imposes limitations on the advancement of relevant applications, particularly for NeRF-Based pointwise query and rendering methods, making it challenging to extend the 3D generation framework to practical projects. What is more, 3D content generation based on the vanilla 3D Gaussian is susceptible to the model getting trapped in local extreme points, leading to artifacts such as burrs, floaters, or proliferative elements in 3D models. Therefore, this paper focuses on using text prompts for 3D content generation and explores possibilities to overcome these limitations.

In summary, our main contributions are as follows:
\begin{itemize}
\item We propose a novel text-to-3D framework based on the Gaussian splatting rendering pipeline and the score function, Langevin dynamics diffusion model. GaussianDiffusion significantly speeds up the rendering process and is able to produce the most realistic appearance currently achievable in text-to-3D tasks.

\item We introduce structured noise from various viewpoints for the first time, aiming to tackle the challenge of maintaining multi-view geometric consistency, such as the problem of multi-faceted structures, through a structured Gaussian noise injection approach. Furthermore, we propose an effective method for generating structured noise based on the Gaussian diffusion process.

\item To address the inherent contradiction between precise Gaussian graphics modeling and the instability observed in 2D diffusion models across multiple views, we introduce a variational Gaussian Splatting model to mitigate the risk of the 3D Gaussian model converging to local minima, which may cause artifacts like floaters, burrs or proliferative elements.
\end{itemize}

%% file: sec/2_related.tex
\begin{figure}[t]
    \centering
    \includegraphics[width=1\linewidth]{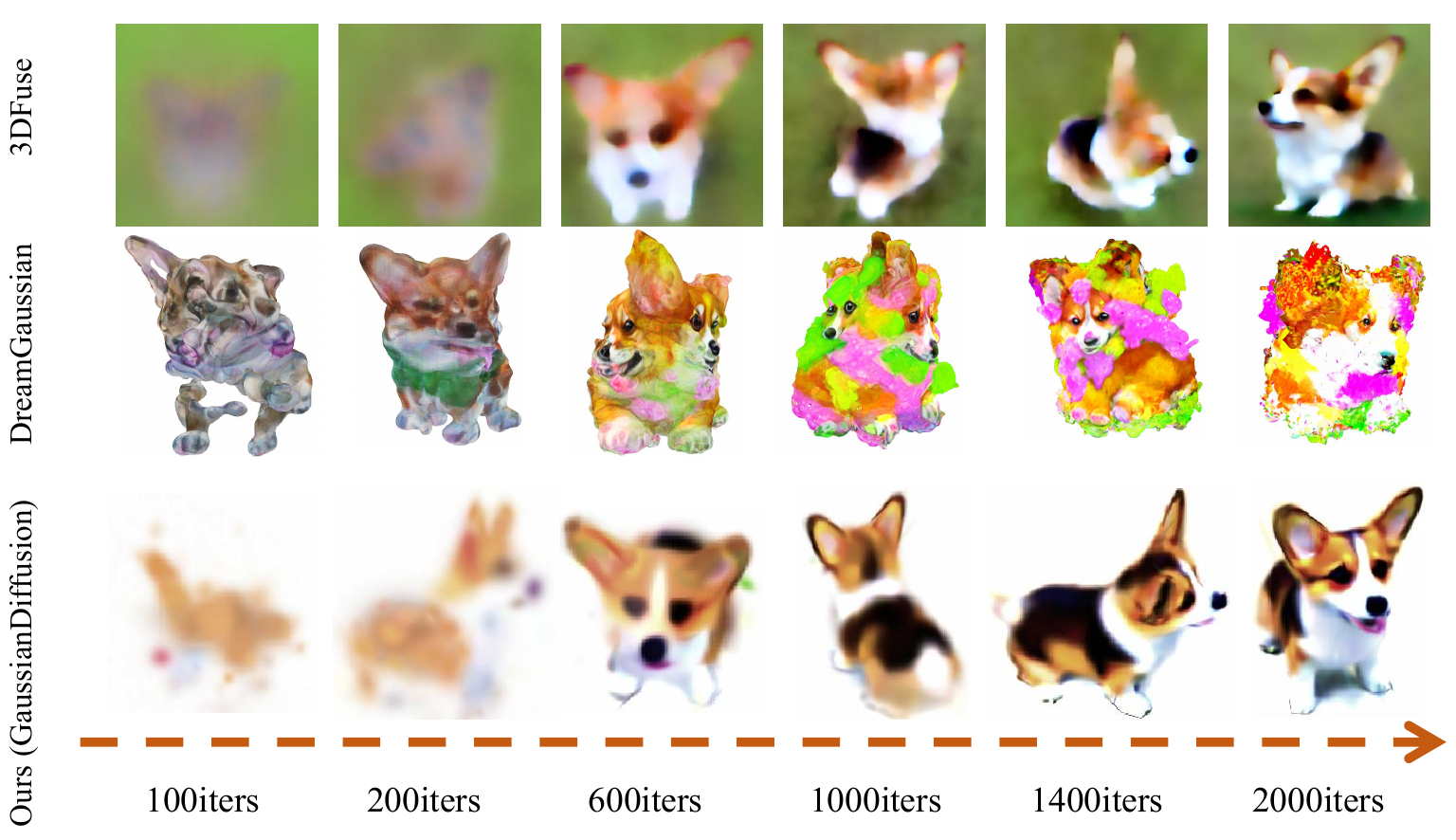}
    \vspace{-1em}
    \caption{Comparision of convergence speed.}
    \label{fig3}
    \vspace{-1em}
\end{figure}

\section{Related Works}
\subsection{Diffusion models}
Diffusion models~\cite{ho2020denoising,song2020score,song2019generative}, have recently gained widespread attention in the field of 2D image generation owing to their stability, versatility, and scalability. In the text-to-image domain, CLIP~\cite{radford2021learning} was first introduced to guide text-to-image generation  by GLIDE~\cite{nichol2021glide}. Subsequently, there have been many developments in text-to-image frameworks that offer higher resolutions and fidelity, such as Imagen~\cite{saharia2022photorealistic}, DALL-E2~\cite{ramesh2022hierarchical}, and Stable-Diffusion~\cite{rombach2022high}. The emergence of numerous 2D text-to-image models has laid the groundwork for 3D generation. This allows 3D models to build upon mature 2D generation models, utilizing them as teacher models for distillation-based learning.

\begin{figure*}[t]
    \centering
    \includegraphics[width=1\linewidth]{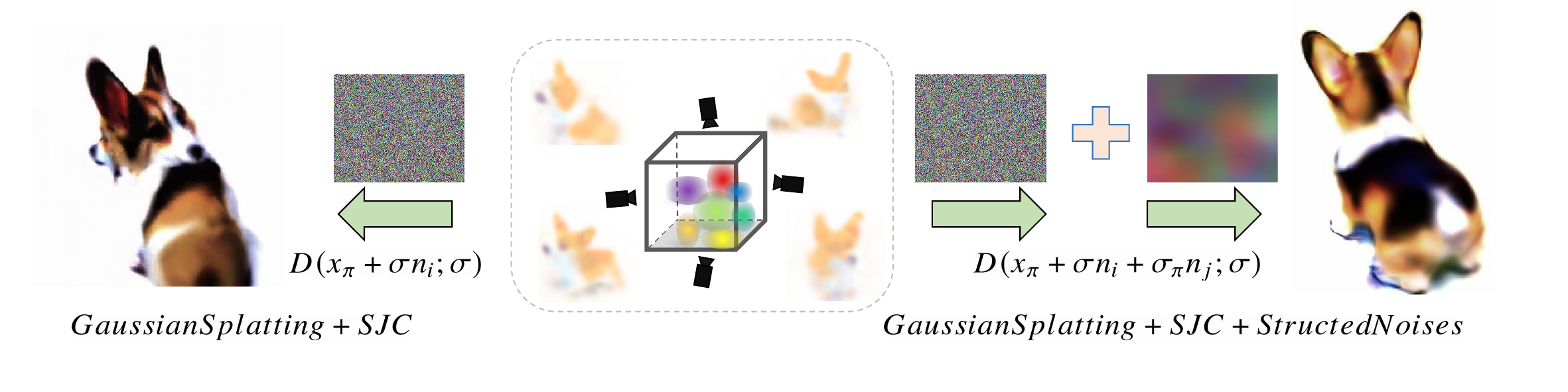}
    \vspace{-2em}
    \caption{Structured Noise. The left portion in the figure represents the SJC method. It involves adding noise to $x_{\pi}$ to gradually transform it into a standard normal distribution $N(0,I)$, and evaluate $D(x_{\pi} + \sigma n_i; \sigma)$ through diffusion model. The right portion corresponds to our structured noise approach, which generates additional $N(0,1)$ distributions related to both pose and pixel position from the same noise source. This establishes inherent noise constraints between images generated from different viewpoints, addressing the multi-view consistency problem.}
    \label{fig2}
    \vspace{-1em}
\end{figure*}

\subsection{3D Gaussian Splatting}
3D Gaussian splatting~\cite{kerbl20233d} provides a promising and efficient approach for 3D scene representation. It employs a collection of 3D Gaussian spheres to characterize spatial scenes, with each sphere carrying information about position, scale, color, opacity, and rotation. These spheres are projected into 2D based on camera poses and subsequently blended using $\alpha$-compositing according to their distances to the screen. Through differentiable rendering and gradient-based optimization, the model approximates the rendered images to match the provided ground truth. The unique representation offered by 3D Gaussian splatting not only enables a continuous portrayal of 3D scenes but also seamlessly integrates with traditional rendering pipelines in a discreet form. Thanks to its versatile representation, it has already sparked significant advancements in various directions within NeRF-based methodologies~\cite{mildenhall2021nerf,yu2021pixelnerf,barron2021mip,yu2021plenoctrees,muller2022instant,sucar2021imap,zhu2022nice,kong2023vmap,pumarola2021d,chen2023mobilenerf,peng2021neural,prokudin2021smplpix}.

\subsection{Image-to-3D generation}
The vast amount of image data~\cite{deng2009imagenet,lin2014microsoft} available holds immense potential for single-image-to-3D content generation tasks. At the same time, the maturity of 2D diffusion models~\cite{ho2020denoising,song2020score,song2019generative} has brought new opportunities to excellent frameworks in single-image 3D generation~\cite{chan2022efficient,liu2023zero,qian2023magic123,tang2023make,liu2023one,shi2023zero123pp}. However, generating 3D content from a single image limits the generation and imagination capabilities of computers. Therefore, adopting a text-to-3D approach aligns better with the human-computer interaction model.

\subsection{Text-to-3D generation}
Dreamfusion~\cite{poole2022dreamfusion} pioneered the use of score distillation sampling (SDS) to learn 3D scenes from frozen 2D diffusion models. SJC~\cite{wang2023score} addressed the Out-of-Distribution (OOD) problem between standard normal inputs of 2D diffusion models and 3D rendered images by introducing secondary noise on 3D rendered images through a perturbation process. 3DFuse~\cite{seo2023let} serves as the baseline for our work, guiding the 3D model generation direction in multiple stages, from coarse to fine, while using view-specific depth as a condition and employing ControlNet~\cite{zhang2023adding} to guide the generation direction of the Diffusion model. TANGO~\cite{lei2022tango} transfers the appearance style of a given 3D shape according to a text prompt in a photorealistic manner. However, the method requires a 3D model as input. DreamGaussian~\cite{tang2023dreamgaussian} is our concurrent work, which is based on SDS~\cite{poole2022dreamfusion}, while we draw inspiration from SJC~\cite{wang2023score}, and apply score function and Langevin sampling method~\cite{song2019generative} to recover structure knowledge from Gaussian noise. Our work, GaussianDiffusion, applies Gaussian splatting across the entire spectrum of 3D generation process. Latest text-to-3D works~\cite{tsalicoglou2023textmesh,yu2023points,zhu2023hifa,li2023focaldreamer,wang2023prolificdreamer,huang2023dreamtime,chen2023fantasia3d} produce realistic, multi-view consistent object geometry and color from a given text prompt, unfortunately, NeRF-based generation is time-consuming, and cannot meet industrial needs.

\subsection{Multiface Problem}
The previous methods, such as incorporating viewpoint features~\cite{shi2023mvdream,li2023sweetdreamer,liu2023syncdreamer,qian2023magic123}, required fine-tuning diffusion models on Objaverse~\cite{deitke2023objaverse} data, which involved supervised training. However, due to Objaverse's style influencing 2D diffusion models, fine-tuning tends to result in 3D appearances with an artistic style. Moreover, due to neural network forgetting, this approach severely diminishes the original 2D diffusion model's generalization performance. On the other hand, our structured noise method, akin to SJC~\cite{wang2023score}, doesn't require fine-tuning and doesn't alter the distribution of the original 2D diffusion model. It preserves its generalization capability and realism. Our structured noise is rendered from random Gaussian spheres, combined with the target scene, and then recovered through denoising to produce the 3D scene. This process is a form of self-supervised self-constraint training within the same distribution, shown in Figure ~\ref{fig5}, ensuring the preservation of both generalization capability and realism, making it straightforward and effective.

%% file: sec/3_method.tex
\section{Method}
\subsection{3D Gaussian Splatting for SJC}
The current state-of-the-art baseline for 3D reconstruction is the 3D Gaussian splatting ~\cite{kerbl20233d}, known for its fast rendering and the discretized representation of 3D scenes, significantly facilitating 3D generation and editing. Each Gaussian in the scene is defined by multiple parameters encapsulated in $\theta = \{\boldsymbol{z},\boldsymbol{s},\boldsymbol{q},\alpha,\boldsymbol{c} \}$, where the position $\boldsymbol{z}\in \mathbb{R}^3$, a scaling factor $\boldsymbol{s}\in \mathbb{R}^3$, opacity $\alpha \in \mathbb{R}$, rotation quaternion $\boldsymbol{q} \in  \mathbb{R}^4$, and color feature $\boldsymbol{c} \in  \mathbb{R}^3$ are considered. 

Our objective is to model and sample from the distribution $p(\theta)$, to generate new 3D content. Let $p_{\sigma}(\boldsymbol{x})$ denote the data distribution perturbed by Gaussian noise of standard deviation $\sigma$, As discussed in ~\cite{chan2022efficient,song2020improved}, the denoising score can be approximated as follows, where D is the denoiser.

\begin{equation}
  \nabla_x \log p_{\sigma}(\boldsymbol{x}) \approx \frac{D(\boldsymbol{x};\sigma)-\boldsymbol{x}}{\sigma^2}
  \label{eq:1}
\end{equation}
SJC~\cite{wang2023score} assumes that  the probability density of 3D asset $\theta$ is proportional to the expected probability densities of its multiview 2D image renderings $\boldsymbol{x}_{\pi}$ over camera poses $\pi$:
\begin{equation}
    p_{\sigma}(\theta) \propto \mathbb{E}_{\pi}[p_{\sigma}(\boldsymbol{x}_{\pi}(\theta))]
  \label{eq:2}
\end{equation}
$P_{\sigma}(\theta)$ characterizes the 3D distribution of $\theta$ with the inclusion of a 3D noise perturbation $\sigma$. On the right side of the equation, perturbing with the same 2D Gaussian distribution would inevitably result in confusion from multiple viewpoints. Thus, $\sigma$ needs to incorporate a view-dependent condition $\pi$. Our motivation stems from the concept of perturbing data with noise projected from a common source across various viewpoints, rather than applying 2D noise sampled from the same distribution to generate images. Therefore, Equation~\ref{eq:2} can be refined as:
\begin{equation}
    p_{\sigma}(\theta) \propto \mathbb{E}_{\pi}[p_{\sigma_{\pi}}(\boldsymbol{x}_{\pi}(\theta))]
  \label{eq:3}
\end{equation}
Where $p_{\sigma_{\pi}}$ denotes the data distribution perturbed by Gaussian noise $\sigma_{\pi}$ projected from a common noise source with viewpoint $\pi$. So the lower bound can be rewritten as:
\begin{equation}
    \log p_{\sigma}(\theta) = \log[\mathbb{E}(p_{\sigma_{\pi}}(\boldsymbol{x}_{\pi}))] - log Z
  \label{eq:4}
\end{equation}
where $Z = \int \mathbb{E}_{\pi}[p_{\sigma_{\pi}}(x_{\pi}(\theta))]d\theta$ denotes the normalization constant. So according to the chain rule~\cite{wang2023score}:

\begin{align}
\nabla_{\theta} \log\tilde{p}_\sigma(\theta) &= \mathbb{E}_{\pi} \left[ \nabla_{\theta} \log p_{\sigma_{\pi}}(\boldsymbol{x}_\pi) \right ] \\
\frac{\partial \log\tilde{p}_\sigma(\theta)}{\partial \theta} &= \mathbb{E}_{\pi} \left[ \frac{\partial\log p_{\sigma_{\pi}}(\boldsymbol{x}_\pi)}{\partial\boldsymbol{x}_\pi}  \cdot \frac{\partial\boldsymbol{x}_\pi}{\partial\theta} \right ] \\ 
\label{eq:3D_score}
\underbrace{\nabla_{\theta}\log\tilde{p}_\sigma(\theta)}_{\text{3D score}} 
&= \mathbb{E}_{\pi} [ ~ 
\underbrace{\nabla_{\boldsymbol{x}_\pi}\log p_{\sigma_{\pi}}(\boldsymbol{x}_\pi)}_{\text{2D score; pretrained}} 
\cdot 
\underbrace{ \mathcal{J}_\pi \vphantom{\nabla_{\boldsymbol{x}_\pi}\log p(\boldsymbol{x}_\pi)} }_{\text{renderer Jacobian}} ].
\end{align}

\begin{figure}[t]
    \vspace{-0.5em}
    \centering
    \includegraphics[width=1\linewidth]{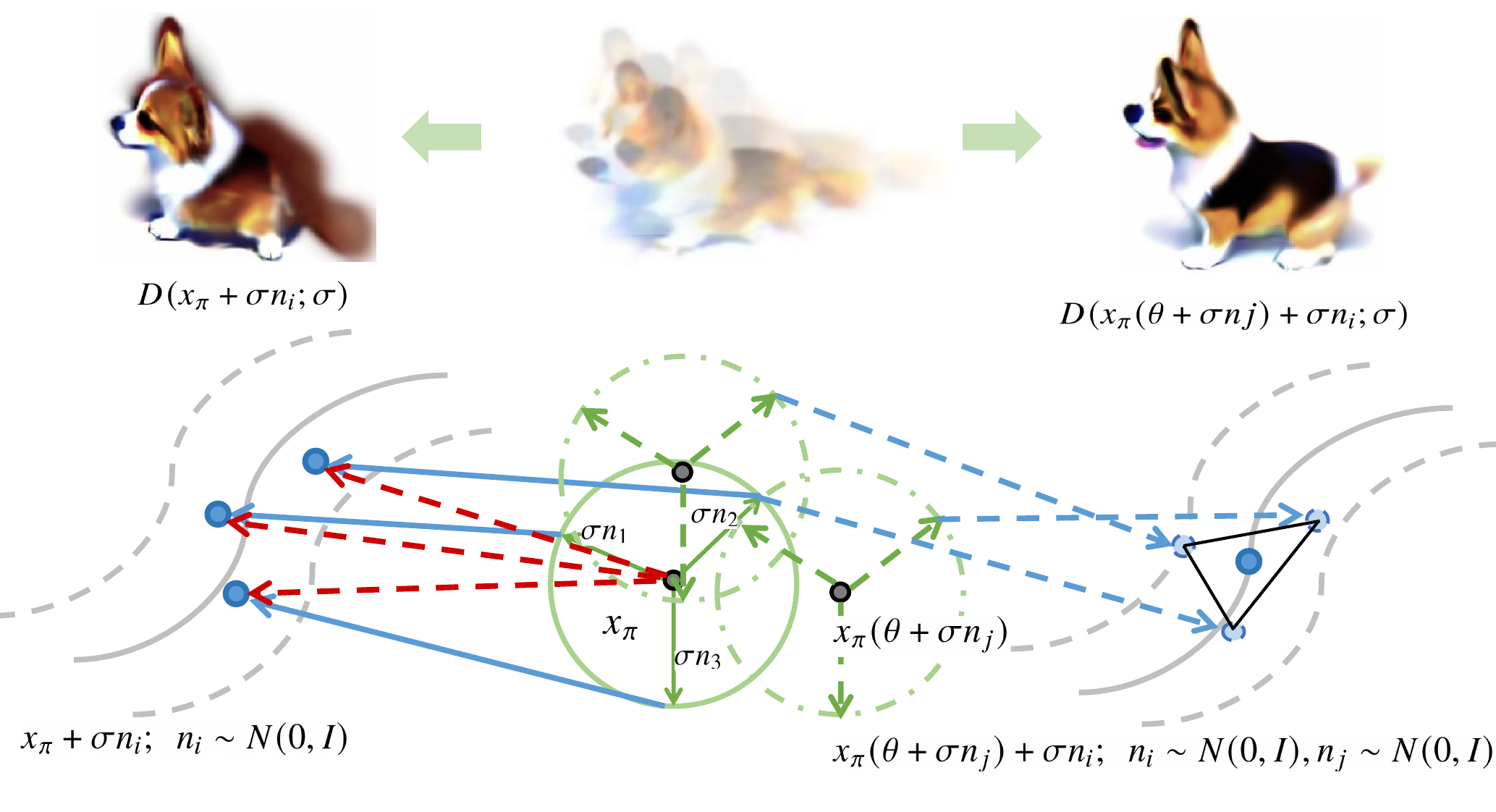}
    \vspace{-1em}
    \caption{Variational Gaussian Splatting. The left portion is the SJC~\cite{wang2023score} method, which involves adding noise to $x_{\pi}$ to gradually transform it into a standard normal distribution $N(0, I)$, and then evaluate $D(x_{\pi} + \sigma n_i; \sigma)$ through diffusion model. On the right, leveraging the variational Gaussian splatting method involves pre-designing a Gaussian model for the parameters $\theta$. During the gradient backward, the gradient is propagated to the mean, while the variance retains the noise level introduced by the diffusion model. The objective is to learn a distribution that more accurately conforms to the correct parameter space by introducing slight variations within a defined range. The distribution points on the triangle are determined by jittering, and then the mean of the distribution is taken as the value for forward inference.}
    \label{fig4}
    \vspace{-0.5em}
\end{figure}

So, the current challenge revolves around efficiently constructing viewpoint-related noise $\sigma_{\pi}$ and utilizing it to perturb $p(x_{\pi}(\theta))$, resulting in $p_{\sigma_{\pi}}(x_{\pi}(\theta))$.

\begin{figure*}[t]
    \centering
    \includegraphics[width=1\linewidth]{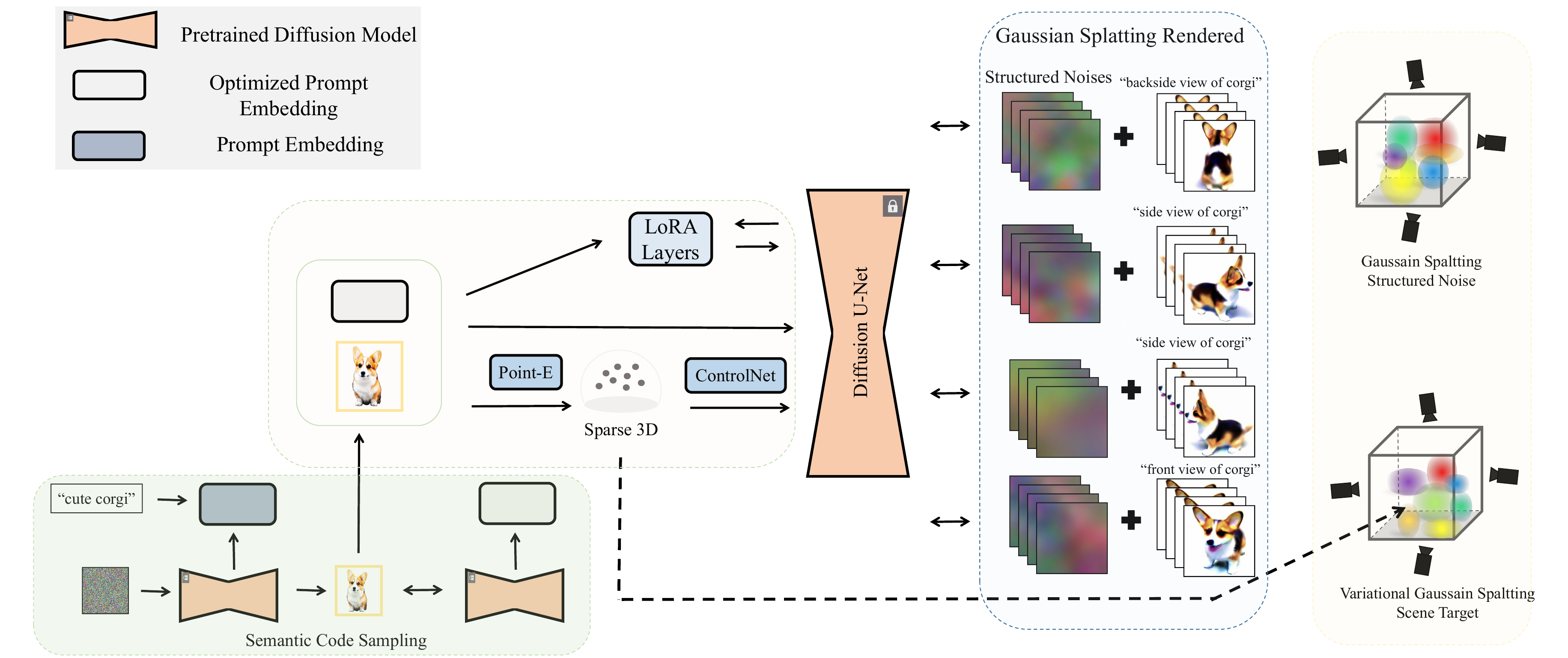}
    \vspace{-1em}
    \caption{GaussianDiffusion Framework. We apply Semantic Code Sampling module~\cite{seo2023let} to restrict the entire 3D scene to a singular semantic identity. An optimized image, derived from Semantic Code Sampling, generates a sparse point cloud using Point-E~\cite{nichol2022point}. This point cloud is subsequently pose-projected into a depth map, functioning as a constraint for ControlNet~\cite{zhang2023adding}. Concurrently, LoRA~\cite{hu2021lora} is deployed for additional optimization for fine-tuning of the diffusion model. The sparse point cloud produced by Point-E acts as the initial input to Gaussian Splatting~\cite{kerbl20233d}. Leveraging SDS~\cite{poole2022dreamfusion}, the gradient of the diffusion model is conveyed to Gaussian Splatting. In order to address challenges related to multi-view consistency and the presence of artifacts such as floaters burrs or proliferative elements, we introduce Structured Noise and the Variational Gaussian Splatting method to produce realistic 3D appearance.}
    \label{fig5}
    \vspace{-0.5em}
\end{figure*}

\subsection{3D Noise Generation and Projection}
For the sake of ensuring positive semi-definiteness, the Gaussian Splatting method is crafted with physically meaningful covariances:
\begin{equation}
    \Sigma = Rss^TR^T
  \label{eq:7}
\end{equation}
Where $s$ represents the scale matrix, $R$ represents the rotation matrix. In the computation, $s$ is used to denote the scale of a 3D vector along the three axes and a quaternion $q$ is used to represent the rotation matrix $R$. Given the viewpoint transformation $W$, $J$ representing the Jacobian of the affine approximation of the projective transformation, the 2D covariance can be projected as:
\begin{equation}
    \Sigma' = JW\Sigma W^TJ^T
  \label{eq:8}
\end{equation}
After projecting the 3D Gaussian ellipsoid onto a 2D plane, it is represented as a bivariate Gaussian distribution on the current ellipse. For a given pixel $U(u_1, v_2)$, the opacities of the Gaussian spheres contributing to that pixel from front to back are $\alpha_1, \alpha_2, ..., \alpha_n$. The final opacities are calculated using Equation~\ref{eq:10}, where $z_{\pi}$ is the projection of the Gaussian sphere position $z$ onto the viewpoint $\pi$.

\begin{equation}
    \alpha_i'(U) = \alpha_i * e^{-(z_{\pi}-U)\Sigma'(z_{\pi}-U)^T}
  \label{eq:10}
\end{equation}

Through the $\alpha$-blending, the final color is then determined by:
\begin{equation}
\begin{split}
C &= \sum_{i=1}^{n} \alpha'_i c_i \prod_{j=1}^{i-1} (1-\alpha'_j)  \\
 &= \alpha'_1 c_1+(1-\alpha'_1)[\sum_{i=2}^{n} \alpha'_i c_i \prod_{j=2}^{i-1} (1-\alpha'_j)]
  \label{eq:9}
\end{split}
\end{equation}
Let $c_1, c_2, ..., c_n$ follow a normal distribution, and the intermediate term $\alpha'_{n-1}c_{n-1}+(1-\alpha'_{n-1})c_n$ also follows a normal distribution, the iterative process from back to front results in C following a normal distribution.

When the initial 3D noise positions $z$, are uniformly distributed within the range [-0.5, 0.5], and they do not change over time, given pixel coordinates $U(u_1, u_2)$, and a fixed camera viewpoint $\pi$, with fixed $W$ and $J$, then $\alpha' = \alpha * e^{-(z_{\pi}-U)\Sigma'(z_{\pi}-U)^T} = \alpha'(U, W, J, z)$, which signifies that $\alpha_n'$ remains constant and does not change over time, effectively making it a constant. Consequently, the distribution of C follows a linear combination of multiple independent normal distributions and thus is also a normal distribution.

\begin{equation}
\vspace{-2em}
\begin{split}
    \mathbb{E}(C) &= \mathbb{E}[\sum_{i=1}^{n} \alpha_i' c_i \prod_{j=1}^{i-1} (1-\alpha'_j)] \\
    &= \alpha'_1*\mathbb{E}[c_1]+(1-\alpha'_1)*[(\mathbb{E}[\sum_{i=2}^{n} \alpha'_i c_i \prod_{j=2}^{i-1})] \\
    &= 0
  \label{eq:12}
\end{split}
\end{equation}

\begin{align}
\begin{split}
    \text{Var}(C) &= \text{Var}[\sum_{i=1}^{n} \alpha_i' c_i \prod_{j=1}^{i-1} (1-\alpha'_j)] \\
    &= \sum_{i=1}^{n} \alpha_i'^2 \prod_{j=1}^{i-1}(1-\alpha'_j)^2 \\
  \label{eq:13}
\end{split}
\end{align}
So, for a given pixel position, C can be represented as a standard normal distribution. That is $P(C|U,\pi) = N(0, \text{Var}(C))$ can be standardized to $N(0, I)$, allowing for noise to be added to the novel view synthesized by Gaussian Splatting. This noise is then used in the subsequent diffusion noise-adding and denoising process, the Gaussian diffusion process. This noise is associated with the viewpoint $\pi$ and does not affect the numerical stability of $N(0,I)$ noises added by SJC, as shown in Figure ~\ref{fig2}

\begin{figure*}[t]
    \centering
    \includegraphics[width=1\linewidth]{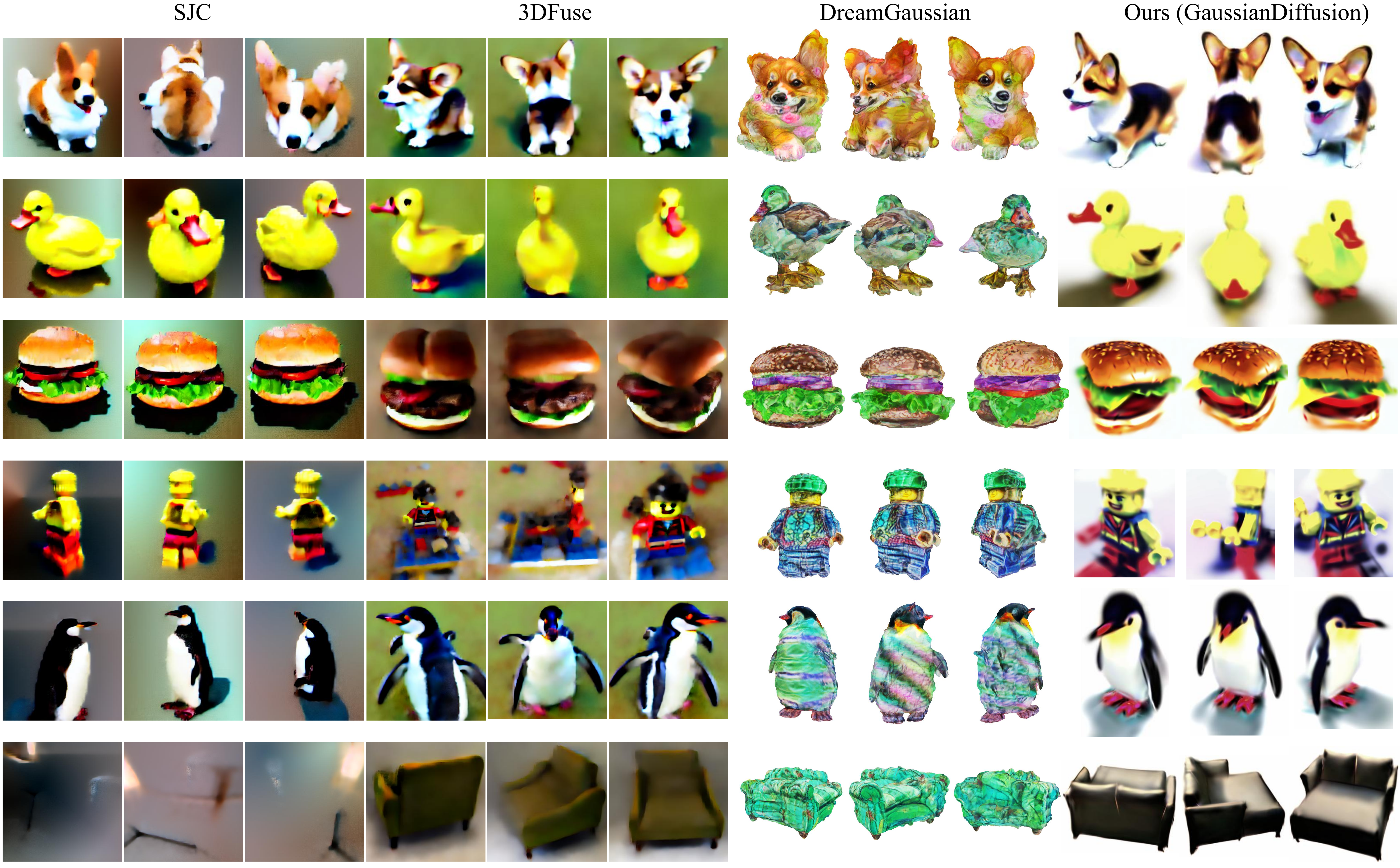}
    \caption{Comparison between SJC, 3DFuse, DreamGaussian, and our GaussianDiffusion in text-to-3D, given text prompts 'a corgi,' 'a yellow duck,' 'a hamburger', 'a Lego figure', 'a penguin', and 'a sofa'.}
    \label{fig6}
    \vspace{-1em}
\end{figure*}

\subsection{Variational Gaussian Spaltting}
There is an inherent contradiction between precise 3D Gaussian Splatting geometric modeling and the inconsistency of multiple views in 2D diffusion model. Beyond the introduction of structured noise, our objective is to fundamentally resolve the imbalance between the two. We formulate the Variational Gaussian Splatting model to propagate the gradient to the distribution of $\theta$, intending to learn the distribution of $\theta$ rather than fixed values. Specifically, it perturbs position $z$ and scale $s$ in the original Gaussian Splatting to facilitate the learned model's transition from a blurry to a precise state and from coarse to fine details.

Modeling the distribution of $z$ and $s$ as a Gaussian model, with the mean of $z$ learned by Gaussian Splatting. The variance, the range of perturbations should synchronize with the noise level in the diffusion model. As mentioned in the SMLD ~\cite{song2019generative}, in regions of low data density, score matching may not have enough evidence to estimate score functions accurately, due to the lack of data samples. When the noise level added by the diffusion model is high, it implies that there is less gradient information in $X_0$, the original picture, so at this time, more perturbations should be added to the original $z$ and $s$ to obtain a blurrier 3D model. When the noise level added by the diffusion model is lower, it means that there is a more effective gradient flow in $X_0$, so correspondingly, less noise should be added to Gaussian Splatting. Guiding the model from coarse to fine, in the rough stage, the perturbations can have looser constraints, which can better balance the multi-view geometric consistency.

For Equation ~\ref{eq:3D_score}, variational Gaussian splatting increases the convergence domain and has the ability to escape local minima. The difference between $p_{\sigma}(\theta)$ and $p_{\sigma}(X)$ is that $p_{\sigma}(X)$ directly expresses the distribution of X with the addition of $\sigma$ perturbations, while in the expression of $p_{\sigma}(\theta)$, there is actually a latent variable. $p_{\sigma}(g(\theta))$ represents the space given by the parameter $\theta$, and through the mapping of the function $g$, it maps $\theta$ to the 3D space, where $g(\theta)$ represents the 3D scene space. Perturbing $\theta$, the coverage domain of $g(\theta)$ expands, and when performing distillation learning on the 2D diffusion model, it helps to escape local minima and avoid artifacts such as burrs, floaters, or proliferative elements.
We design the parameters $\theta'$ to follow a Gaussian distribution $N(\theta, \sigma)$, where $\sigma$ is a variance at the same level as the noise added by the diffusion model:
\begin{equation}
\theta' = \theta + \sigma*\mathcal{N}(0,I)
\label{eq15}
\end{equation}
Equation ~\ref{eq:17} indicates that the gradient with respect to $\theta'$ is equivalent to the gradient with respect to $\theta$. This allows for the introduction of noise perturbation without altering the gradient flow, facilitating the transition from solving for fixed values to solving for the parameter model distribution.

\begin{align}
\frac{\partial \log\tilde{p}_\sigma(\theta)}{\partial \theta'} 
&= \mathbb{E}_{\pi} \left[ \frac{\partial\log p_{\sigma_{\pi}}(\boldsymbol{x}_\pi)}{\partial\boldsymbol{x}_\pi}  \cdot \frac{\partial\boldsymbol{x}_\pi}{\partial\theta'} \right ] \\
&= \mathbb{E}_{\pi}[ \frac{\partial\log p_{\sigma_{\pi}}(\boldsymbol{x}_\pi)}{\partial\boldsymbol{x}_\pi}  \cdot \frac{\partial\boldsymbol{x}_\pi}{\partial\theta'} \cdot \frac{\partial\theta'}{\partial\theta} ] \\
&= \frac{\partial \log\tilde{p}_\sigma(\theta)}{\partial \theta}
\label{eq:17}
\end{align}

\begin{figure}[t]
    \centering
    \includegraphics[width=1\linewidth]{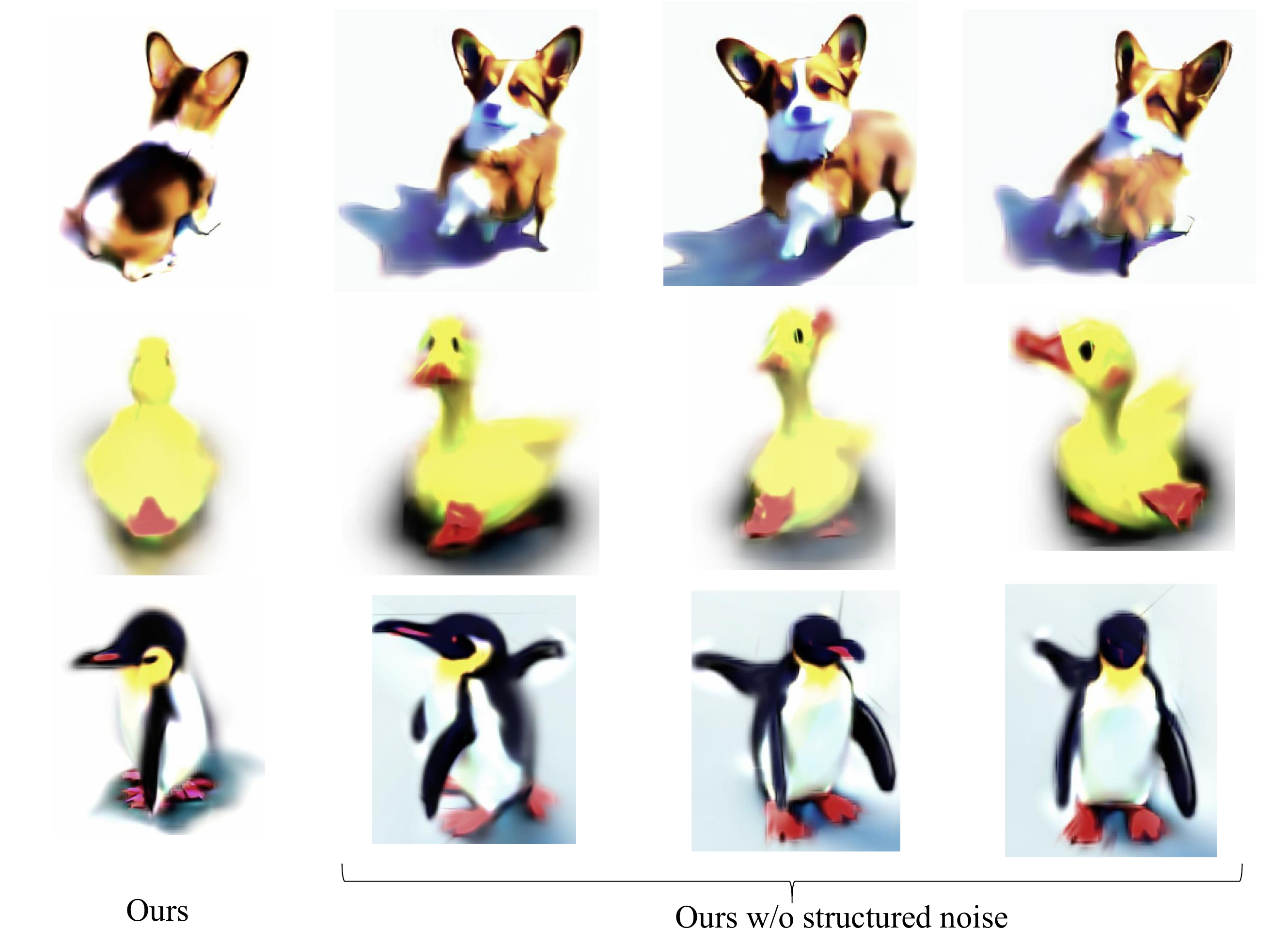}
    \vspace{-1em}
    \caption{Ablation study. GaussianDiffusion without structured noise.}
    \label{fig7}
    \vspace{-0.5em}
\end{figure}

The visualization can be seen in Figure ~\ref{fig4}. On the left, the idea of Computing PAAS~\cite{wang2023score} on 2D renderings, denoted as $x_{\pi}$, is proposed by SJC~\cite{wang2023score}. It involves adding $\sigma n_i$ to the center $x_{\pi}$ to make it approach $\mathcal{N}(0,I)$, addressing the out-of-distribution (OOD) problem, and then evaluate $D(x_{\pi} + \sigma n_i; \sigma)$ through diffusion model. However, this may lead to the generation of artifacts in 3D Gaussian Spaltting such as floaters, burrs, or proliferative elements, as illustrated on the left side of the figure. On the right, building upon this idea, we introduce perturbations to the parameter $\theta$, resulting in additional dashed circles after perturbing $x_{\pi}$. These circles correspond to multiple blurred corgis. Variational Gaussian splatting method involves pre-designing a Gaussian model for the parameters $\theta$. During the gradient backward, the gradient is propagated to the mean, while the variance retains the noise level introduced by the diffusion model.
Through a more relaxed convergence constraint, we ensure that the optimization does not get stuck in a local optimum as shown in the top-left figure, but instead converges to a global optimal solution, as illustrated in the top-right figure.

%% file: sec/4_experiment.tex
\section{Experiment}

\subsection{Overall Pipeline}

As shown in Figure ~\ref{fig5}, we introduce the Semantic Code Sampling module within 3DFuse~\cite{seo2023let} to address the challenge of text prompt ambiguity. In this method, an image is initially generated using the provided text prompt, after which the prompt embedding is optimized based on the resultant image. Subsequently, we apply LoRA~\cite{hu2021lora} adaptation to maintain semantic information and fine-tune the Diffusion model. To integrate 3D awareness into pre-trained 2D diffusion models, we employ Point-E~\cite{nichol2022point} to generate a sparse point cloud from a single image. This point cloud is then projected to derive a depth map utilizing the specified camera pose $\pi$. Furthermore, we incorporate spatial conditioning controls from the depth map into the text-to-image diffusion models~\cite{zhang2023adding}. Simultaneously, this sparse point cloud serves as the input point cloud for Gaussian Splatting initialization.


We conduct training over 2000 steps for the overall stage, whereas 3DFuse requires 10000 steps for achieving stable results. The 3D Gaussians are initially set with 0.1 opacity and a grey color within a sphere of radius 0.5. Gaussian splatting is performed at a rendering resolution of 512. Random camera poses are sampled at a fixed radius of 1, with a y-axis FOV between 40 and 70 degrees, azimuth in the range of [-180, 180] degrees, and elevation in the range of [-45, 45] degrees. The background is rendered randomly as white. The 3D Gaussians are initialized with 4096 cloud points, derived from the output of Point-E~\cite{nichol2022point}. The cloud is densified every 50 steps after 300 steps and has its opacity reset for 400 steps. This opacity reset guarantees that our final appearance remains free from oversaturation. All experiments are conducted and measured using an NVIDIA 4090 (24GB) GPU.
\subsection{Comparision of Convergence Speed}
The comparison of convergence speeds can be observed in Figure ~\ref{fig3}. The convergence speed of DreamGaussian~\cite{tang2023dreamgaussian} is relatively high, yet it is susceptible to overfitting in later stages, lacks a continuous optimization space, and demonstrates suboptimal multi-view geometric consistency. The 3DFuse~\cite{seo2023let} method, requiring approximately 10,000 iterations for convergence to satisfactory results, still manifests issues such as overexposure and burr artifacts. Similar to fundamental NeRF-based approaches, it also renders downstream tasks at a slower pace. In contrast, our method achieves significantly faster convergence, reaching a satisfactory state around 2,000 iterations while ensuring geometric consistency. It successfully avoids artifacts like floaters burrs or proliferative elements, resulting in the generation of the most realistic appearance. Additionally, being rooted in 3D Gaussian methods, it notably enhances rendering speed without a sudden slowdown with increasing pixels. Furthermore, downstream applications seamlessly integrate with traditional rendering pipelines.

\subsection{Structured Noise Generation Details}
We ingeniously apply Gaussian Splatting  to generate random pixeltwise normal distribution noises. The point cloud is initialized in spherical coordinates with radii ranging from -0.5 to 0.5. Azimuth is randomly selected between -180 and 180 degrees, and elevation between -45 and 45 degrees. The initialization color of the point cloud is sampled from a standard normal distribution. As detailed in the 3D Noise Generation and Projection section, for a given pixel $U(u_1,u_2)$ and a specified camera pose $\pi$, we ensure that the added noise adheres to a standard normal distribution, preserving the numerical stability of the original image. The diffusion model's input noise comprised a combination of structural noise and SJC~\cite{wang2023score} noise, wherein the proportion of structural noise declined gradually from 0.3 to 0.05 over a span of 2000 iterations, as shown in Figure ~\ref{fig2}.


\subsection{Variational Gaussian Splatting}
We apply noise with the same magnitude variance $\sigma$ of the frozen diffusion model and zero mean, to perturb the parameters $z$ and $s$ as the description of Eq ~\ref{eq15}. Based on experimental observations, we found that applying a coefficient of 0.1 to the jitter noise variance yields better results, denoted as $\theta' = \theta + \sigma \cdot N(0, I) \cdot \gamma$, where $\gamma$ equals 0.15. The comparison results between whether add perturbation $\theta$ or not can be seen as Figure \ref{fig4}.

\subsection{Quantitative Comparison}
We reference the geometric consistency for 3D scene generation quantification in 3DFuse~\cite{seo2023let}, which is based on COLMAP~\cite{schonberger2016structure}. The underlying principle is that COLMAP optimizes camera poses based on the multi-view geometric consistency of the 3D surface. So, we uniformly sample 100 camera poses from the hemisphere coordinates, all facing the center of the sphere with the same radius and elevation angle. We then render 100 images using these poses and predict the poses between adjacent images using COLMAP~\cite{schonberger2016structure}. The variance of the predicted poses is used as a measure of 3D consistency. High variance indicates inaccuracies in pose prediction, implying 3D geometric inconsistency.

\begin{figure}[t]
    \centering
    \includegraphics[width=1\linewidth]{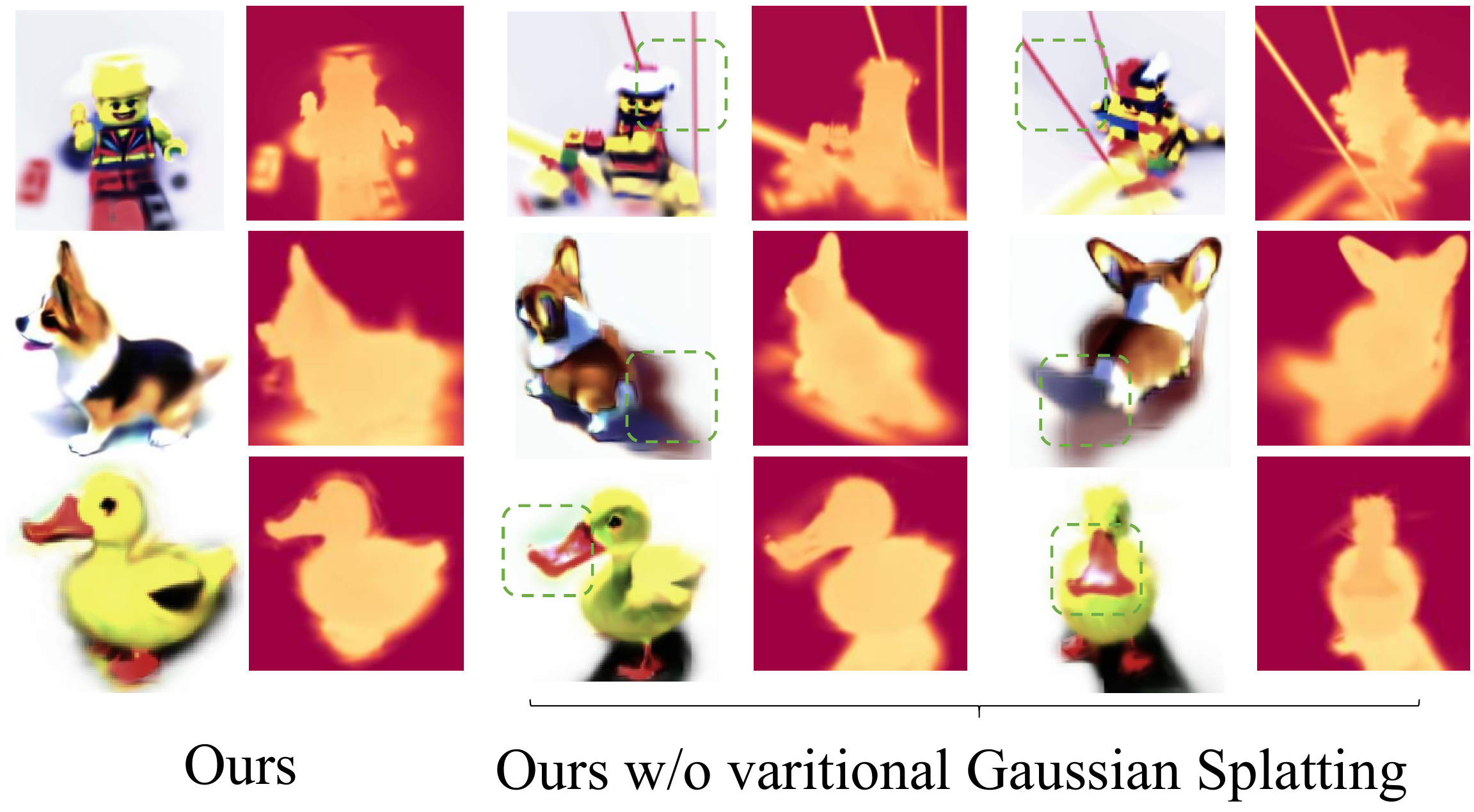}
    \caption{Ablation study. GaussianDiffusion without varitional Gaussian splatting.}
    \label{fig8}
    \vspace{-1em}
\end{figure}

In Table ~\ref{tab1}, we compared the mean variance scores of 50 generated 3D scenes. Experimental data demonstrates that our method significantly outperforms SJC~\cite{wang2023score}, 3DFuse~\cite{seo2023let}, and DreamGaussian~\cite{tang2023dreamgaussian}, providing strong evidence that our approach exhibits superior geometric consistency in text-to-3D tasks.

\begin{table}[!htbp]
	\centering
    \normalsize
 	\resizebox{0.48\textwidth}{!}{
		\begin{tabular}{cccc}
			\toprule
	Method	 & Variance $\downarrow$ & Train Iters & Training Time\\ 
    \midrule
			SJC~\cite{wang2023score}  & 0.081 &10000 & 13.8min\\
			3DFuse~\cite{seo2023let} & 0.053 & 10000 & 20.3min\\
			DreamGaussian~\cite{tang2023dreamgaussian} & 0.106 &600 & 3.1min\\	
			GaussianDiffusion(Ours) & 0.021 & 2000 & 5.5min\\
			\bottomrule
	\end{tabular}
 }
	\caption{Quantitative evaluation of mean variance in 50 generated 3D scenes based on COLMAP~\cite{schonberger2016structure} method proposed by 3DFuse~\cite{seo2023let}.}
    \label{tab1}
    \vspace{-1em}
\end{table}

\subsection{Qualitative Comparision}
As depicted in Figure ~\ref{fig6}, we present a comparative analysis of our experimental results, wherein all methods are meticulously fine-tuned for optimal convergence. The SJC~\cite{wang2023score} and 3DFuse~\cite{seo2023let} methodologies undergo training for 10,000 iterations, while DreamGaussian~\cite{tang2023dreamgaussian} undergoes 600 iterations, and our method is trained for 2,000 iterations. Experimental validation reveals that SJC's earlier proposed method exhibits shortcomings in both multi-view geometric consistency and appearance. Although 3DFuse demonstrates an acceptable appearance and a certain level of geometric consistency, there are still issues with multi-view consistency, as seen in the penguin, and it contends with issues such as overexposure, floaters, burrs, or proliferative elements in appearance. DreamGaussian~\cite{tang2023dreamgaussian} achieves swift convergence, yet its limited subsequent optimization space and poor multi-view consistency are drawbacks. Furthermore, its texture optimization operates as a post-processing method independently of the overall pipeline. In contrast, our method, incorporating structured noises and variational Gaussian Splatting, consistently attains multi-view geometric consistency, yielding realistic and visually appealing outcomes.

\subsection{Ablation Study}
As depicted in Figure ~\ref{fig2} and ~\ref{fig7},  after removing the technology of structured noise from multiple viewpoints, it is evident that there are issues with multi-view geometric consistency and multi-faceted structures. Additionally, the variance in the quantitative analysis shows a sharp decline, as illustrated in Table ~\ref{tab2}.

\begin{table}[!htbp]
	\centering
    \normalsize
 	\resizebox{0.48\textwidth}{!}{
		\begin{tabular}{ccc}
			\toprule
	Method	 & Variance $\downarrow$ & Train Iters \\
    \midrule
			GaussianDiffusion w/o Structured Noise  & 0.056 &2000 \\
            GaussianDiffusion w/o Variational Gaussian   & 0.033 &2000 \\
			GaussianDiffusion(Ours) & 0.021 & 2000 \\
			\bottomrule
	\end{tabular}
 }
	\caption{Ablation study. We compared the effects after removing structured noise and variational Gaussian Splatting with the results obtained from the complete version.}
    \label{tab2}
    \vspace{-1em}
\end{table}

As illustrated in Figure ~\ref{fig4} and ~\ref{fig8}, replacing the variational Gaussian Splatting technique with vanilla Gaussian Splatting results in numerous floaters, burrs, or proliferative elements. The quantitative analysis shows a slight decrease in variance, as indicated in Table ~\ref{tab2}.

%% file: sec/5_conclusion.tex
\section{Discussion and Conclusion}
This paper introduces a 3D content generation framework based on Gaussian Splatting, significantly accelerating rendering speed while achieving the most realistic appearance to date in text-to-3D tasks. By incorporating structured noise from multiple viewpoints, we address the challenge of multi-view geometric inconsistency. What is more, the variational Gaussian Splatting technique enhances the generated appearance by mitigating artifacts like floaters, burrs proliferative elements. While the current results demonstrate improved realism compared to prior methods, the utilization of variational Gaussian introduces some degree of blurriness and haze. Consequently, our forthcoming research endeavors aim to rectify and enhance this aspect.